\documentclass[runningheads]{llncs}
\usepackage{graphicx}
\usepackage{booktabs}
\usepackage{multirow}   
\usepackage{subcaption}
\usepackage{xcolor}

\begin{document}
\title{Visualizing the embedding space to explain the effect of knowledge distillation}
\titlerunning{Visualizing knowledge distillation}
%\thanks{Supported by Korea University}}
%
\author{Hyun Seung Lee\inst{1}\orcidID{0000-0002-4088-3632} \newline
Christian Wallraven\inst{2,*}\orcidID{0000-0002-2604-9115}}

\authorrunning{H.Lee and C.Wallraven}

%\email{wallraven@korea.ac.kr}}
\institute{Department of Artificial Intelligence, Korea University, Seoul, Korea \inst{1} \newline \email{hslrock@korea.ac.kr}\newline Department of Artificial Intelligence \& Department of Brain and Cognitive Engineering, Korea University, Seoul, Korea \inst{2} \newline \email{wallraven@korea.ac.kr}
}  

\maketitle              % typeset the header of the contribution

\begin{abstract}    

   Recent research has found that knowledge distillation can be effective in reducing the size of a network and in increasing generalization. A pre-trained, large teacher network, for example, was shown to be able to bootstrap a student model that eventually outperforms the teacher  in a limited label environment. Despite these advances, it still is relatively unclear \emph{why} this method works, that is, what the resulting student model does 'better'. To address this issue, here, we utilize two non-linear, low-dimensional embedding methods (t-SNE and IVIS) to visualize representation spaces of different layers in a network. We perform a set of extensive experiments with different architecture parameters and distillation methods. The resulting visualizations and metrics clearly show that distillation guides the network to find a more compact representation space for higher accuracy already in earlier layers compared to its non-distilled version.
   
\end{abstract}

\section{Introduction}

The field of image recognition has rapidly developed with convolutional neural networks that allow for efficient computation of filters learned from large amounts of data.  Researchers were able to stack multiple convolutional layers and form a "deep" network \cite{he2016deep,simonyan2014very} to increase performance. Empirically, it was found that \textit{deeper} networks have higher accuracy in many benchmark datasets. At the same time, however, \textit{deeper} networks require more computational resources. To address this issue, researchers have developed different methods to compress the network without significant loss of performance, including pruning and quantization. \cite{han2015deep}.  

Here, we focus on knowledge distillation \cite{hinton2015distilling}, another method designed to \textit{compress} the network by \textit{transferring} the knowledge through soft targets from a trained network. Based on this approach, additional distillation schemes have been developed (such as attention-based, quantized, and multi-teacher) that further improve efficiency \cite{gou2021knowledge,komodakis2017paying,polino2018model,liu2020}. Recently, SimCLR2 \cite{chen2020big} was proposed - an architecture, which is effective in limited data training through interactions between a well-trained, large teacher network and another, smaller student network. Surprisingly, the distillation method showed that the student model could even \textit{outperform} the teacher in certain cases. Therefore, knowledge distillation has become one of the commonly-used techniques in transfer learning and other, similar application areas. Attempts to \textit{explain} the mechanism behind the knowledge distillation, however, are non-trivial since distillation relies on one black-box model's output to train another, new black-box model. To our knowledge, it is still largely unclear \emph{how} distilled and undistilled networks differ in their representation of the classification problem.

This paper aims to investigate the effect of distillation in terms of accuracy, correlations, and - most importantly - representation space. To obtain the latter, we compare two different non-linear, low-dimensional embedding methods (t-SNE \cite{van2008visualizing}, and the neural-network-based Ivis \cite{szubert2019}). We use different metrics from these embeddings to measure the effect of distillation and also visualize the representation spaces for qualitative exploration of the networks' generalization ability.

This paper has three main key contributions: First, we examine the effect of the knowledge distillation methods in different few-shot learning scenarios. Second, we show that distillation better captures the mutual information among class labels. Third, we use the embedding methods to visualize the representation spaces to explore the effect of distillation.

\section{Related Work}

\subsection{Dimensionality Reduction}
    In a deep network, the latent space is of "\textit{high dimensionality}". However, this high-dimensionality is often a problem in explaining the network since its decision pattern is not easy to visualize. Therefore, researchers have developed different methods to reduce the number of dimensions while trying to preserve as much of the original information as possible. 
    
    Principal component analysis (PCA) is one such reduction method based on the correlation within  dimensions. However, it is a linear mapping making it not suitable for data containing non-linearities. t-SNE \cite{van2008visualizing} was developed to specifically produce a non-linear, low-dimensional visualization using stochastic neighbor embedding. Specifically, this method computes the probability distribution of multiple embeddings assigning higher probability to more similar embeddings and vice versa. After this, it uses the calculated probability distributions to determine the most similar distribution in the target dimensionality. One common issue with t-SNE is that it has numerous different parameters that influence the result and has high computational cost in for a large number of input dimensions \cite{wattenberg2016use}. Also, t-SNE by design often creates clustered data points even with unclustered random data input (see below). Despite this, t-SNE has been a popular methods to visualize the representation space of neural networks (see \cite{bengio2012practical} for an early recommendation of using t-SNE and \cite{seifert2017visualizations} for an early review). In practice, however, research has mostly employed t-SNE either on the raw, small-scale input or on comparatively low-dimensional parts of the network, such as vectorized words \cite{aljalbout1801clustering,zhu2017uncovering} given its computational cost for higher-dimensional inputs (but see \cite{yu2014visualizing}).
    
    Recently, Benjamin et al. \cite{szubert2019} developed an alternative reduction model called Ivis. This is a neural network-based approach that learns a parametric mapping from a higher, input dimensionality to a lower, target dimensionality by minimizing the \textit{triplet loss} of a Siamese network. The networks use anchor, positive, and negative comparison pairs from the input data sampled using a nearest neighbor algorithm. Through the shared Siamese parallel network, the framework reduces the input to the target dimensionality. Later, the network computes the triplet loss between samples and fits its parameters towards the best embedding generator - see \ref{sec:methods}. Overall, Ivis showed \textit{higher} effectiveness in capturing the global data structures compared to other methods \cite{szubert2019}. 
  
    \begin{figure}[ht]
      \centering
    \begin{minipage}[t]{0.3\textwidth}
      \includegraphics[width=\linewidth]{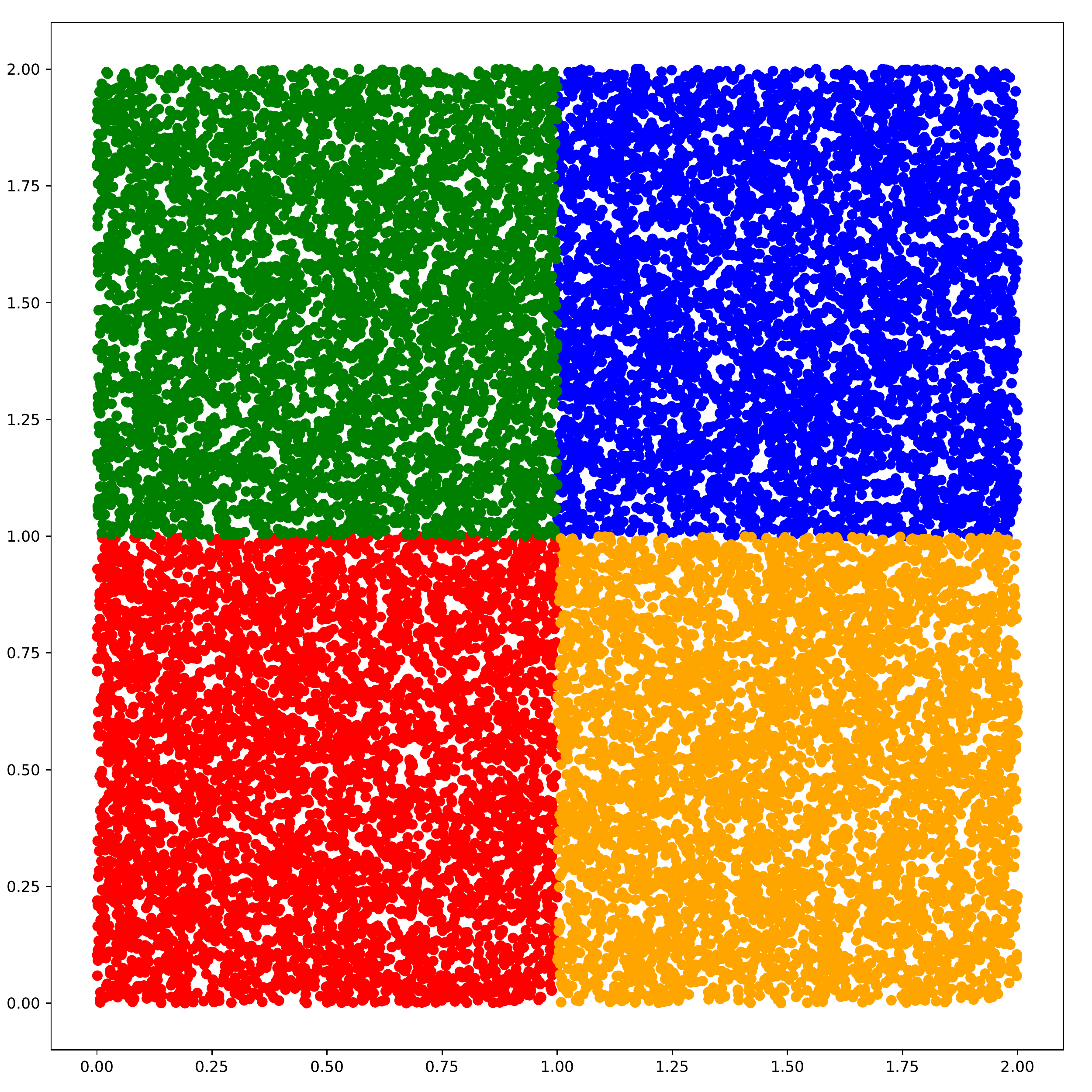}
      \subcaption{Random Data}\label{fig:random_noise}
    \end{minipage}\hspace{0.3cm}
    \begin{minipage}[t]{0.3\textwidth}
      \includegraphics[width=\linewidth]{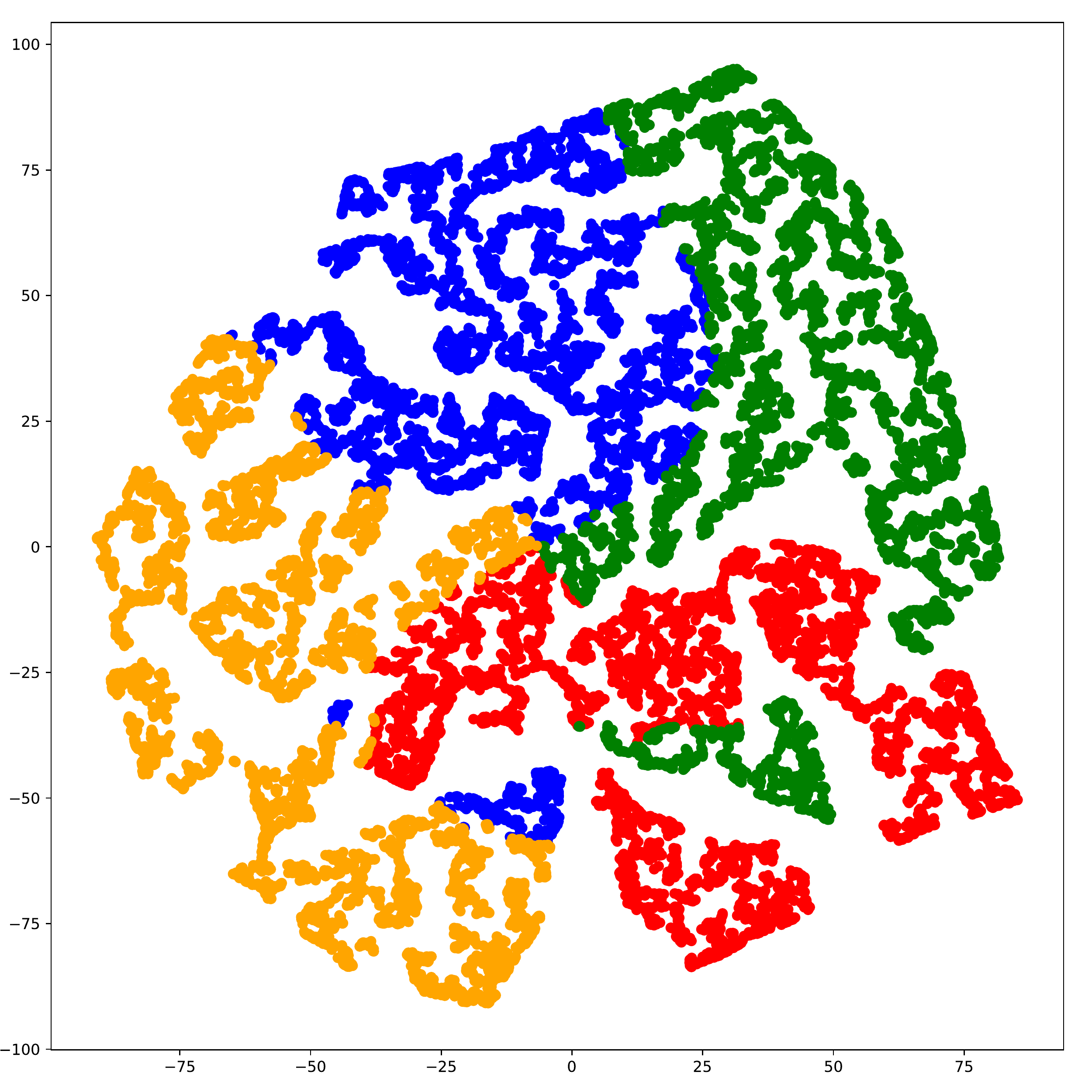}
      \subcaption{t-SNE}
      \label{fig:model2}
    \end{minipage}\hspace{0.5cm}
    \begin{minipage}[t]{0.3\textwidth}
      \includegraphics[width=\linewidth]{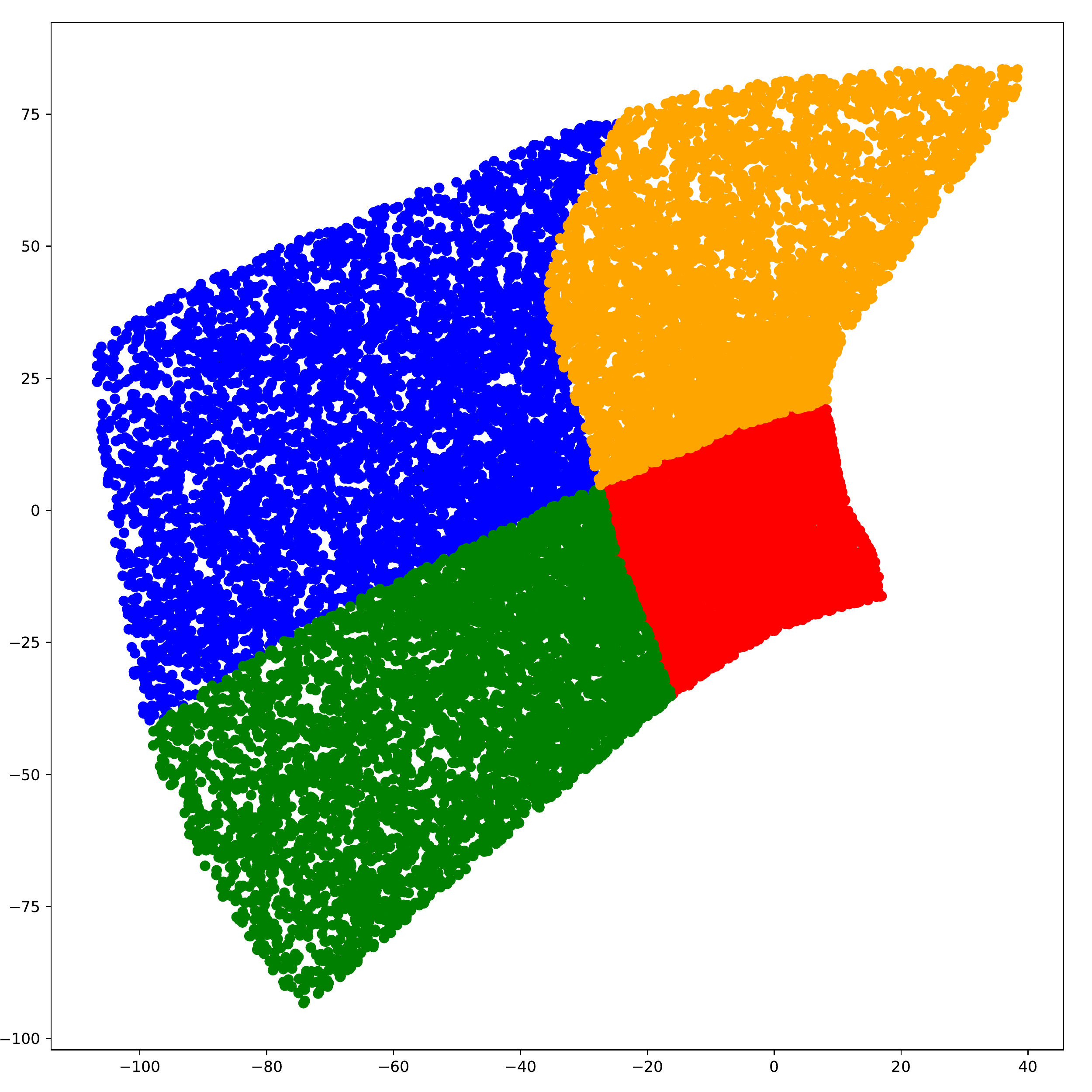}
      \subcaption{Ivis method}
      \label{fig:model3}
    \end{minipage}
    \caption{2D uniform noise (a) projected to 15 dimensions and re-embedded with b) t-SNE and c) Ivis.}
    \label{fig:three graphs}
    \end{figure}

    Fig.\ref{fig:three graphs} illustrates a dimensionality reduction result for uniform random noise (color-labelled into four 2D quadrants), comparing t-SNE and Ivis for illustration. We projected this 2D random noise input into a higher dimension ($\Re^2 \longrightarrow \Re^{15}$) by performing both linear and non-linear mappings ($x+y,x-y,x^2,y^2,sin(x+y),e^x,x^3,y^3$ ...) and then applied the two methods. As the figure shows, Ivis is better able to preserve the global information compared to the cluster-biased t-SNE method.
 
\begin{figure}[ht]

\centering
\begin{subfigure}[b]{0.4\textwidth}
\centering
  \includegraphics[width=\textwidth]{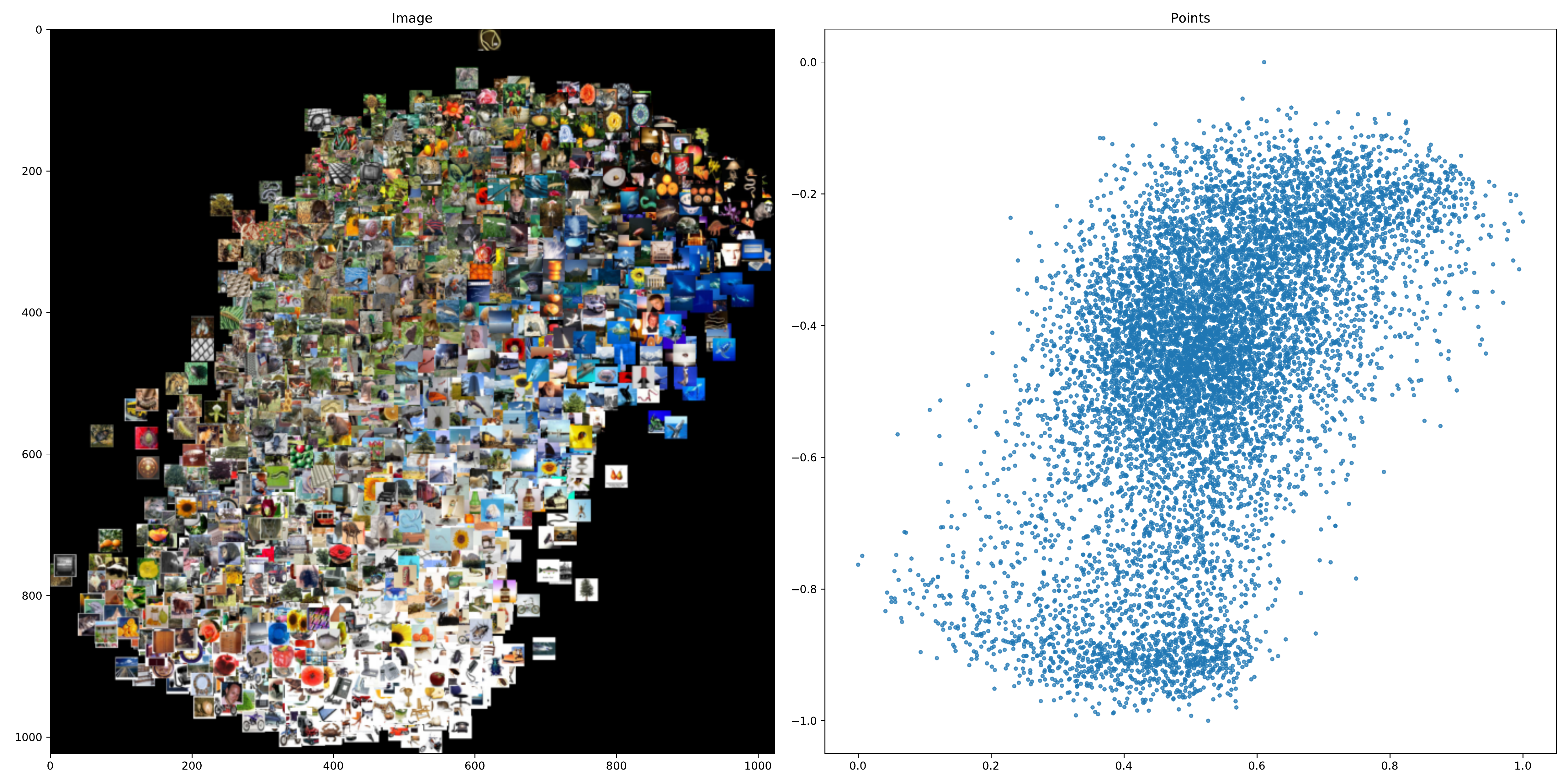}
  \captionof{figure}{Ivis method}

\end{subfigure}
\begin{subfigure}[b]{0.4\textwidth}
\centering
  \includegraphics[width=\textwidth]{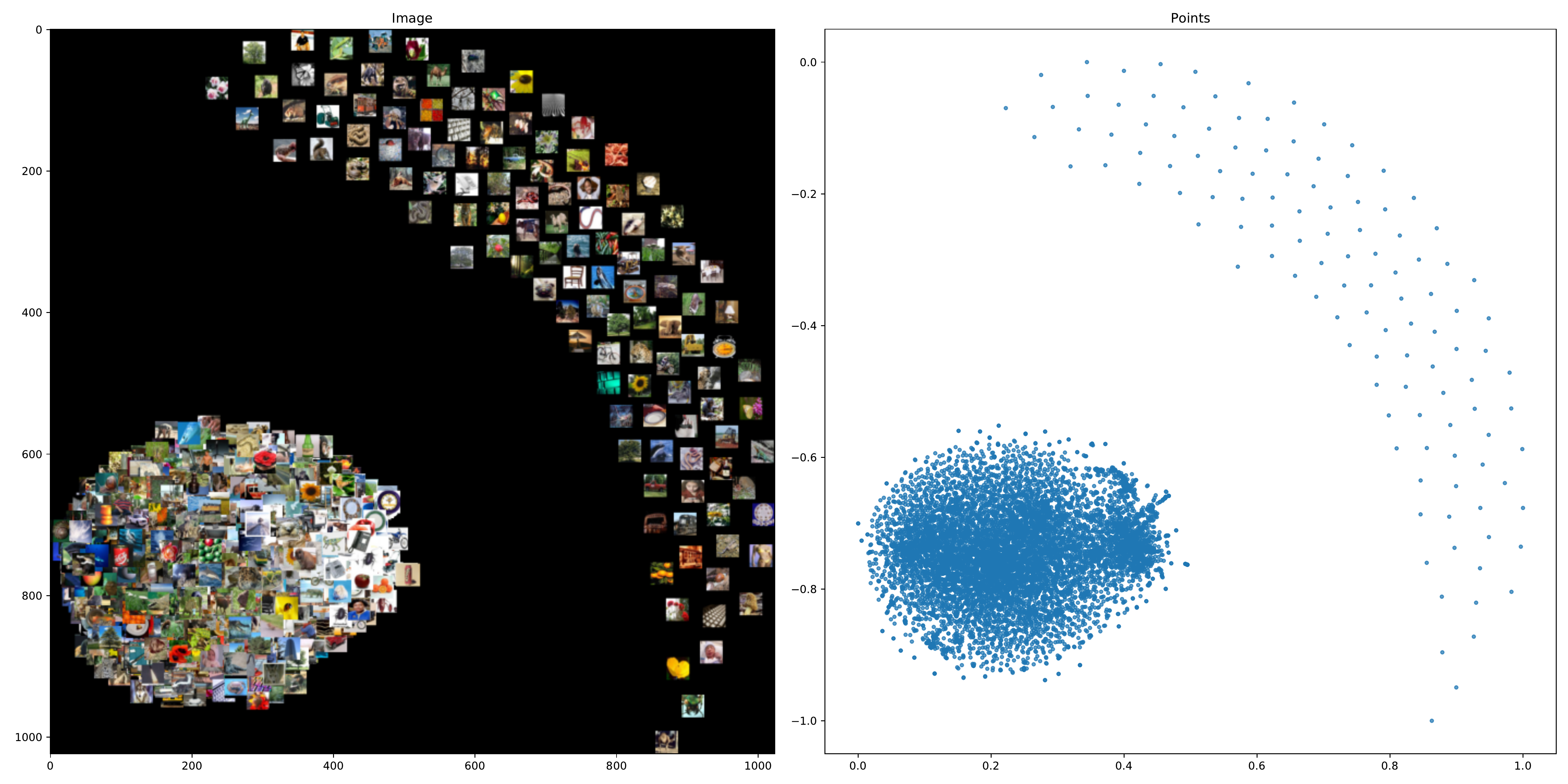}
  \captionof{figure}{t-SNE method}

\end{subfigure}

\caption{Two-dimensional embedding of the CIFAR-100 dataset, comparing Ivis and t-SNE.}
        \label{fig:visualization}
        
\end{figure}

     One of the advantages of using the structure-preserving Ivis method while explaining the network's behavior is that humans can directly observe and understand the network's representation spaces. Fig \ref{fig:visualization}. compares two-dimensional embeddings from the Ivis method and the t-SNE method from an embedding of the popular CIFAR-100 dataset \cite{krizhevsky2009learning} (see below for more details). As it is not possible to determine ground truth, we rely on a preliminary, qualitative analyses first. In line with Fig.\ref{fig:random_noise}, we observe that Ivis produces a smoother distribution of the test dataset of CIFAR-100, compared to t-SNE, which often creates outlier points. At the same time, Ivis has a significantly lower computational budget: 600s to fit 10000 samples with 16384 dimensions, compared to t-SNE's computation time with 3700s at a perplexity of 150 (but see \cite{chan2018t} for GPU-enabled computation of t-SNE, which may decrease its execution time).
    
    The goal of this paper is to compare these two methods to visualize the high-dimensional representation spaces of distilled and undistilled networks. 
  
\subsection{Transfer Learning} 

\textit{Transfer learning} refers to the use of a model (pre-)trained for a certain task in another, second task \cite{zhuang2020comprehensive}. Most current applications use such pre-trained network for weight initialization and subsequent fine-tuning. Canziani et al. \cite{canziani2016analysis}  showed that this approach can achieve higher accuracy compared to random weight initialization. Fine-tuning the network was shown to be especially effective when the network needs to work in a domain with a small sample size. Additional approaches to address such limited sample situations make use of data augmentation, label mixing, and consistency regularization \cite{sohn2020fixmatch,berthelot2019mixmatch,tarvainen2017mean,arazo2020pseudo,rasmus2015semi}. 

Compared to standard transfer learning, which directly uses the pre-trained weights, \textit{knowledge distillation} is another method that is analogous to an interaction between a teacher and their student. Hinton et al. \cite{hinton2015distilling} pioneered this method by using the (soft) outputs from a large teacher network for training a smaller student network (this method bears similarity to label smoothing \cite{muller2019does}). In their experiments, the student's performance showed similar performance compared to the teacher - importantly, follow-up studies showed that distillation actually resulted in \emph{better performing} student networks in a limited label setting in \cite{chen2020big,lin2019progressive,thiagarajan2019distill}.

Recently, the so-called FixMatch\cite{sohn2020fixmatch} approach was proposed that uses self-interaction inside the training model to improve consistency and confidence without a separate teacher model. It does this by using weak and strong augmentations, where weak augmentation flips and shifts input images, whereas strong augmentations consists of randomly-selected unrestricted transformations. For all augmentations, the goal of the final model was to match its outputs on both augmented images. This method has achieved state-of-the-art performance on CIFAR-10,100 and SVHN in limited label settings. One issue with this approach, however, is that the optimal augmentation strategy that does not affect the essential features of input images needs to be hand-crafted for each dataset - knowledge distillation in contrast is a more automated process.

\subsection{Explaining Knowledge Distillation}
Although knowledge distillation created a new field in transfer learning, it remained a black box. According to a recent study by Wang et al. \cite{wang2021knowledge}, knowledge distillation maximizes the mutual information between the teacher and the student as shown through Bayes' rule: by fitting to the general teacher's information, an effective student can be trained. Phuong et al. \cite{phuong2019towards} showed that distillation is equivalent to learning with a favored data distribution, unbiased information, and strong monotonicity. Finally, Cheng et al. \cite{cheng2020explaining} presented a mathematical model trying to quantify the discarded information throughout the layers in the models with and without distillation - they showed that the student seems to discard task-irrelevant information, learns faster, and optimizes with fewer detours.

To our knowledge, however, further \textit{visualizations} of the representation space of the different layers of distilled and undistilled networks in the contexts of limited-label setting and transfer learning have not been explored so far, which is the focus of the present work. Note, that we use the term "limited-label setting" for investigating approaches on how to use knowledge from a large amount of unlabeled data to improve classification when given only a small amount of labeled data - other relevant contexts for our work are the semi-supervised learning or few-shot learning fields.

\section{Methods}\label{sec:methods}

\subsection{Distillation Training}
 This paper uses a \textit{Wide-Resnet} as a student model and a standard \textit{Resnet} as a teacher network \cite{zagoruyko2016wide,he2016deep}. These two network architectures have become "standard" backbone networks used often in smaller-scale classification tasks such as CIFAR-100 \cite{krizhevsky2009learning}. Irwan et al.\cite{bello2021revisiting} showed that a large Resnet remains at the top rank for CIFAR-100 with modification of the training methods, achieving 89.3\% accuracy without any extra data used. For our distillation experiments, we used a Wide-Resnet with a width multiplier of one and a depth scale of 28, denoting this network as\textit{ WideRes-28-1}. For the "larger" Resnet, we used a Resnet-18 that was \textit{pre-trained} on  ImageNet. The ratio of parameters in WideRes-28-1 to Resnet-18 is 3\% with 0.37 million parameters for the former. We added an extra upsampling layer at the top of Resnet-18 for fine-tuned with CIFAR-100 for size-matching.

\begin{figure}

\centering
\begin{subfigure}[b]{0.45\textwidth}
\centering
  \includegraphics[width=\textwidth]{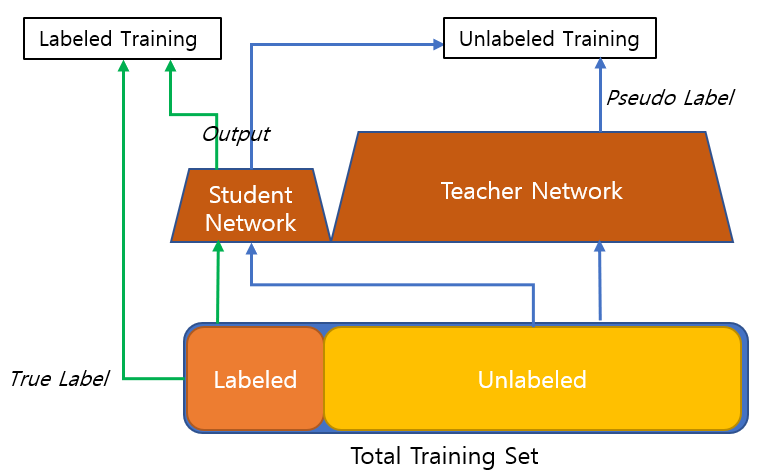}
  \captionof{figure}{Distillation training method}
\end{subfigure}
\hfill
\begin{subfigure}[b]{0.45\textwidth}
\centering
  \includegraphics[width=\textwidth]{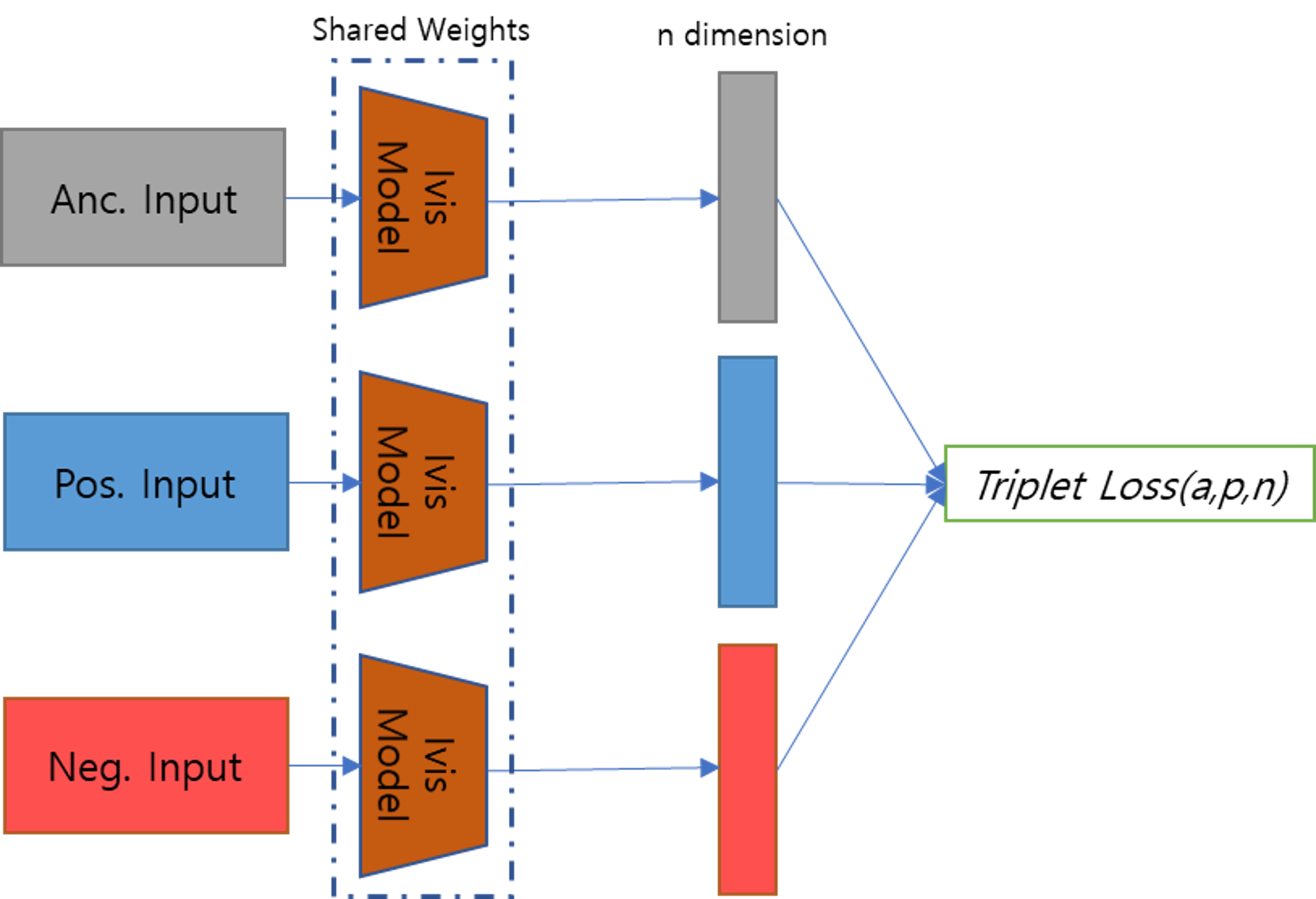}
  \captionof{figure}{Ivis method}
\end{subfigure}
\caption{Training methods used in the experiment}\label{fig: training}

\end{figure}

In experiments on limited sample settings, we extract subsets of CIFAR-100 using balanced sampling. We refer to the labeled and unlabeled data as  $x^l \in R^n$ and $x^{ul}\in R^n$. We initially train without any interaction between the student network and teacher network, where we label the untrained student as $f(x)$ and the teacher as $F(x)$. After the teacher network finishes its fine-tuning on the supplied training set, we use $F(x^{ul})$ to train a distilled student network, $g(x)$. As Fig.\ref{fig: training} shows, we undergo cyclical update of $g(x)$ by first training with $x^l$, then we compute $F(x^{ul})$ for unlabeled data and update $g(x)$'s weights using the loss function, Eq.(\ref{distillation}).
 
    \begin{equation}
        \sum_{x^{ul} \in X^{ul}} KL(softmax(\frac{g(x^{ul})}{T}),softmax(\frac{F(x^{ul})}{T}))
        \label{distillation}
    \end{equation}
This loss decreases the distributional difference between the two networks, with temperature $T$ as a smoothing factor.  
  
\subsection{Ivis Analysis}
 After training, we analyzed the difference between the two models, $f(x)$ and $g(x)$. For our initial analysis, we employed the Ivis method of dimensionality reduction with the test dataset of CIFAR-100 (see below for a comparison with t-SNE).
 
As expressed in Fig.\ref{fig: training}, Ivis selects the anchor, positive, and negative samples from the input data. Then, it decreases each input embedding dimension with size $n$ through three networks with shared parameters. The network then modifies its parameters based on the \textit{triplet loss} between the embeddings: 
  \begin{equation}
    \sum_{i}^N 	[||I(x_i^a)-I(x_i^p)||_2^2-||I(x_i^a)-I(x_i^n)||_2^2+\alpha ]
    \label{triplet}
 \end{equation}
 
As shown in Eq.(\ref{triplet}), the triplet loss is computed using the comparison of an anchor input distance with positive and negative inputs \cite{schroff2015facenet}. The positive and negative samples are selected through nearest-neighbor sampling in an \emph{unsupervised} method, known as \textit{Annoy} index \cite{Bernhardsson} - the closest sample being the positive sample and the furthest being the negative one. The network maximizes the distance between negative and anchor samples and minimizes it between positive and anchor samples. The loss includes $\alpha$ as the threshold for the margin between positive and negative pairs. Thus, we can obtain an embedding with a reduced dimension that keeps its original information by minimizing the triplet loss. This reduced embedding and the converged loss score will be used in our analysis.

\section{Experiments}

    \subsection{Datasets}
        
     We analyzed the effect on the standard classification dataset of CIFAR-100 \cite{krizhevsky2009learning}, which consists of 100 classes with 600 images per class. Originally, the data is split into 500 training and 100 test images per class, but given our limited sample setting, we only used a subset of these for training and testing. Additionally, CIFAR-100 has extra-label information called a 'superclass,' which group the 100 labels more coarsely. The Resnet-18 teacher network was first trained using the ImageNet database \cite{russakovsky2015imagenet}, which is a large dataset containing more than 14 million images in 1000 classes. 
     
     \subsection{Implementation Details}
     
      Both teacher and student networks used \textit{cross-entropy loss}, and for the distillation, $T=4$ was used (lower values did not result in significant effects, whereas too high values led to divergence). We used the SGD optimizer for updating the weights. The initial learning rate was $0.1$ with step-size decay per $60$ epochs with multiplicity of $0.2$. We set the batch size to 64. For augmentations, only random horizontal flips were applied during training. 
      
      For the CIFAR-100 dataset, we tested the distillation with 400, 2500, 5000, 10000, 20000 total labeled data budgets (balanced in each class). For comparison in this limited sample setting, below, we also report results shown in the recent FixMatch work\cite{sohn2020fixmatch} and other related work. FixMatch used a much larger WideResnet-28-8 for their training, so that we also used their framework with a comparably-sized WideResnet-28-1. For all analyses of distillation effects, we report average accuracy and standard deviation across five random folds. 
      
\begin{figure}[ht]

        \centering
        \begin{subfigure}[b]{0.45\textwidth}
        \centering
          \includegraphics[width=\textwidth]{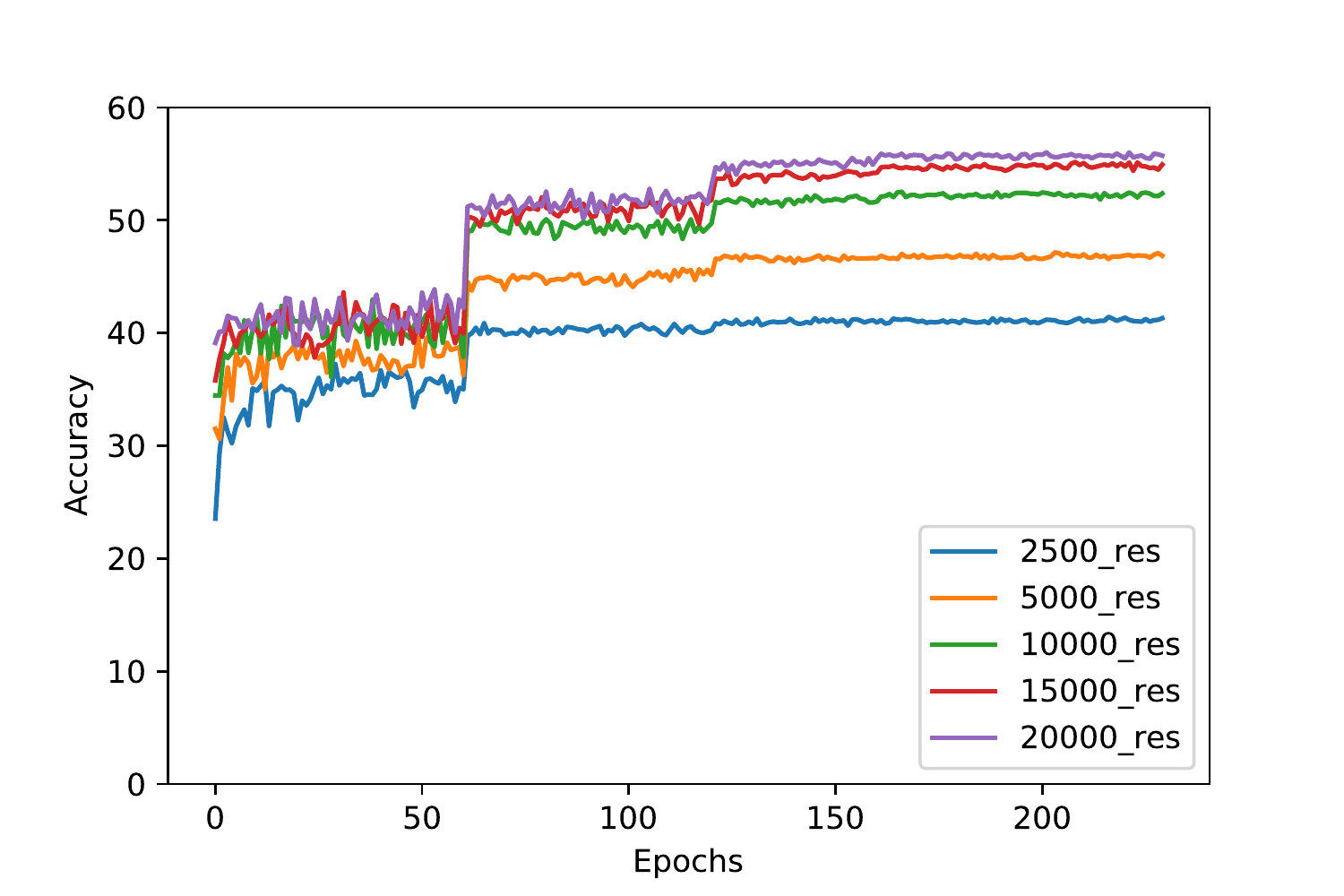} 
          \captionof{figure}{Pretrained Resnet-18}
        \end{subfigure}
        \begin{subfigure}[b]{0.45\textwidth}
        \centering
          \includegraphics[width=\textwidth]{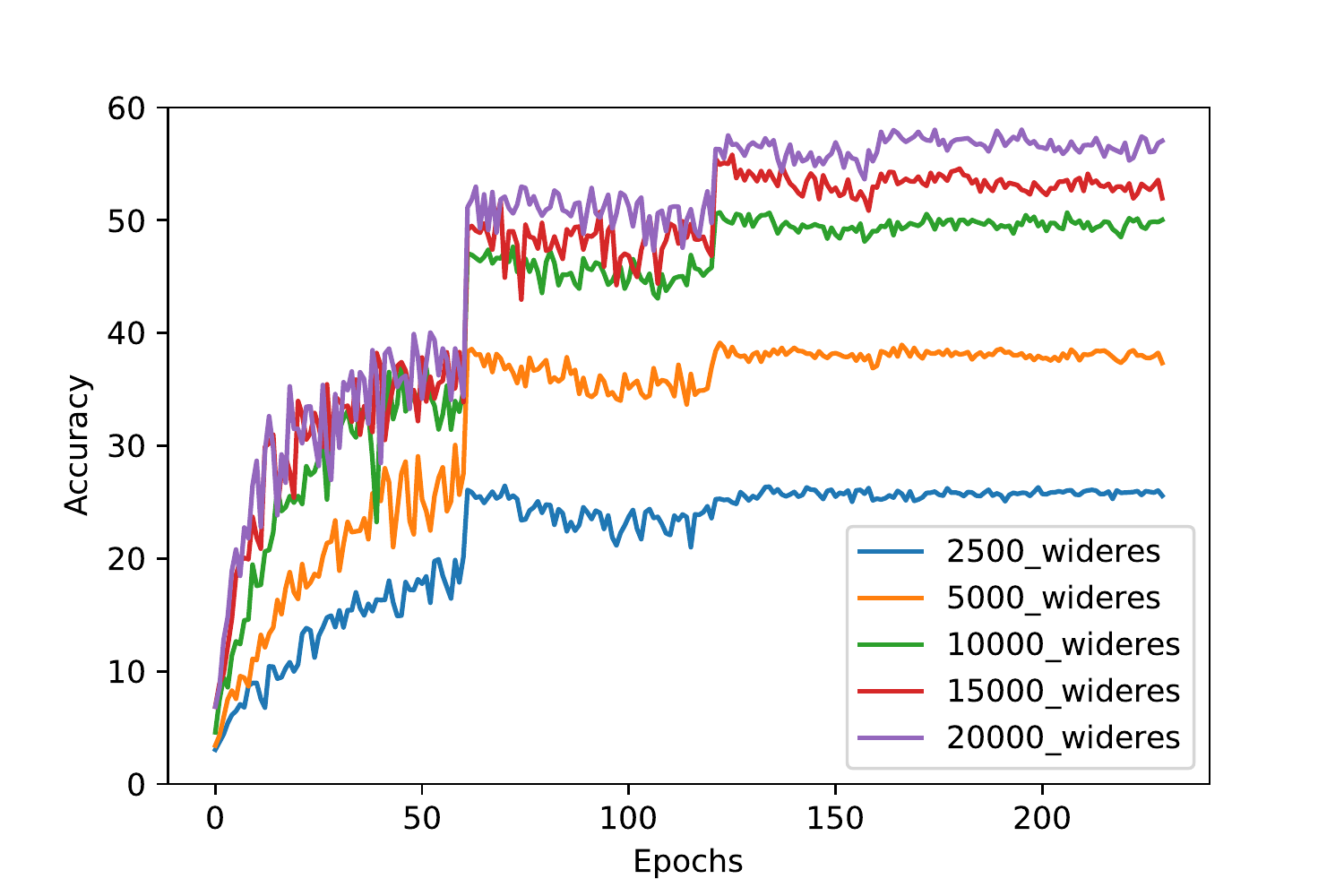}
          \captionof{figure}{WideResnet-28-1}
        \end{subfigure}
        
        \caption{Accuracy comparison for a pre-trained Resnet vs a "vanilla" Wide-Resnet in various, limited sample settings.}
                \label{fig:accuracy_test}
                
        \end{figure}

      Fig.\ref{fig:accuracy_test} shows some of the test accuracies of WideResnet-28-1 and Resnet-18 with different number of labels in CIFAR-100. As visible, a pre-trained Resnet-18 has better performance in the presence of limited labeled data, compared to the untrained WideResnet-28-1. When using 20,000 labels (equivalent to 40\% of the full training set), however, the Wide-Resnet approaches similar to higher performance 
      
%     \begin{figure}[ht]
%
%    \centering
%  
%     \includegraphics[width=0.4\textwidth]{img_comparison.pdf}
%     \captionof{figure}{Zoomed-in comparison of two label budgets for the Resnet vs the Wide-Resnet.}
%     \label{zoom_comparison}
%     
%    \end{figure} 

\section{Results and Discussion}

    \subsection{Classification Accuracy}
Table \ref{table1} presents detailed results for distillation. With more than 20,000 samples, there is no clear effect of distillation with similar performance between distilled and undistilled student networks and the teacher network. This result may represent the capacity limit of the student network. In this context, it is important to note that the pattern of degradation in the performance of the state-of-the-art FixMatch approach is similar to the distilled model with the same (reduced) model size. Overall, the distilled model also showed similar, high robustness to reductions in label budget comparable with the teacher’s network. Similar to other distillation research, the student network can perform better compared to the teacher network in highly-constrained budgets (bold values in Table \ref{table1}).
\begin{table}[ht]
    \caption{Accuracy for CIFAR-100 for different limited label methods. FixMatch \cite{sohn2020fixmatch} performed best overall, but with reduction of the model's backbone to a Wide-Resnet-28-1, performance became similar in most conditions in our experiments (below horizontal line; UD = undistilled, D = distilled).}\label{tab1}

    \centering
    \setlength{\tabcolsep}{5pt}
    \begin{tabular}{lccccc}
        \toprule
        {} &    \multicolumn{5}{c}{CIFAR-100} \\
        \cmidrule{2-6}  
        Method  & 400    & 2500   & 5000   & 10000  & 20000 \\
        \cmidrule(lr){1-1} \cmidrule{2-2} \cmidrule{3-3} \cmidrule{4-4}\cmidrule{5-5}\cmidrule{6-6}
       % \midrule
II-Model\cite{rasmus2015semi}               &- & 42.75$_{\pm0.48}$ & -     & 62.12$_{\pm0.11}$ & -     \\ 
Pseudo-Labeling \cite{tarvainen2017mean}    & - & 42.62$_{\pm0.46}$ & -     & 63.79$_{\pm0.19}$ & -     \\ 
Mean Teacher \cite{sohn2020fixmatch}        & - & 46.09$_{\pm0.57}$ & -     & 64.17$_{\pm0.24}$ & -     \\ 
MixMatch  \cite{berthelot2019mixmatch}      & 32.39$_{\pm1.32}$ & 60.06$_{\pm0.37}$ & -     & 71.69$_{\pm0.33}$ & -     \\      
FixMatch \cite{sohn2020fixmatch}            & 51.15$_{\pm1.75}$ & 71.71$_{\pm0.11}$  & -     & 77.40$_{\pm0.12}$ & -     \\ 
FixMatch (Reduced)                           & 25.90 & 45.63 &  -     & 60.35 &       \\ 
\midrule
WResnet-28-1(UD)    & 5.60$_{\pm0.82}$ & 24.24$_{\pm1.30}$ & 35.77$_{\pm1.67}$ & 48.33$_{\pm0.98}$ & \textbf{57.18}$_{\pm1.94}$ \\ 
WResnet-28-1(D)      & 23.75$_{\pm0.85}$      & \textbf{46.22}$_{\pm0.61}$ & \textbf{50.15}$_{\pm0.80}$ & \textbf{54.24}$_{\pm1.69}$ & \textbf{56.51}$_{\pm1.43}$      \\ 
Resnet-18               & 26.18$_{\pm0.51}$ & 41.48$_{\pm0.77}$ & 47.11$_{\pm0.06}$ & 52.11$_{\pm0.12}$ & \textbf{55.97}$_{\pm0.41}$ \\
        \bottomrule
    \end{tabular}
    \label{table1}
\end{table}

\subsection{Correlation of final outputs}

\begin{figure}[ht]
\centering
  \includegraphics[width=\linewidth,height=4cm]{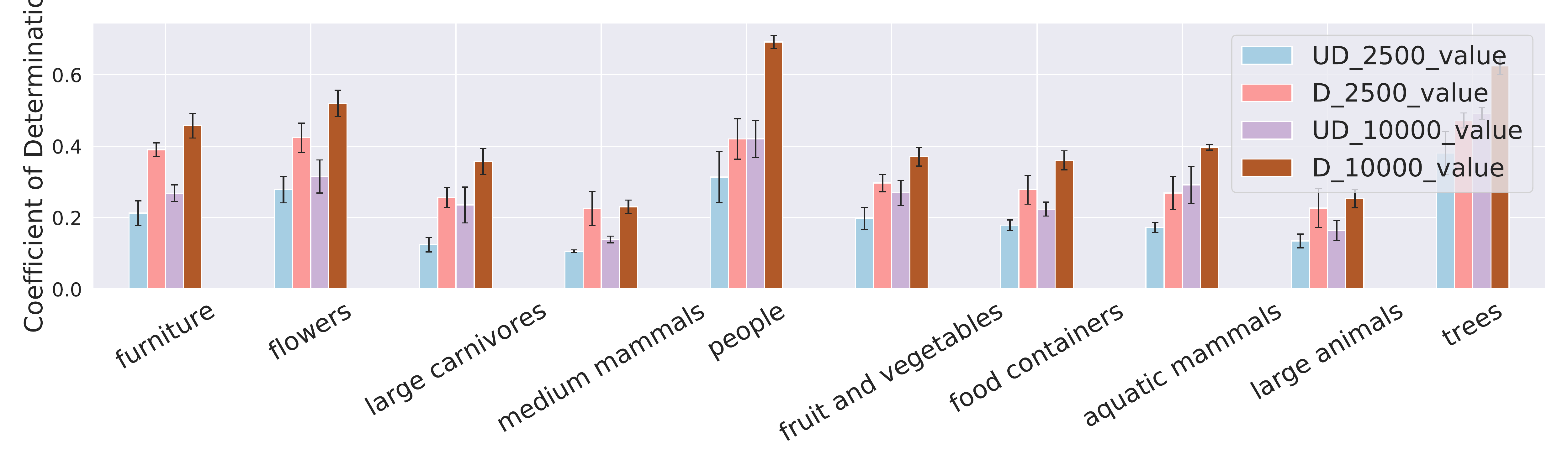}
  \caption{Correlation comparison $(r^2)$ between coarse labels in a single super class}
\label{CorrelationChart}

\end{figure}

    We next explored correlations for the final outputs as represented by the real-valued vote towards each label in CIFAR-100. Fig.\ref{CorrelationChart} shows this correlation, which was computed as the mean correlation value between all labels contained \emph{in the same superclass} for distilled ($\mathrm{D}_{*}$) and undistilled networks ($\mathrm{UD}_{*}$). As can be seen, the distilled model often can represent the inter-class correlation better compared to the undistilled model. This is especially the case when coarse labels inside a superclass have high visual similarity: for example, the people class has a large effect on correlation strength as it contains man, woman, baby, girl, and boy as visually similar categories. For this superclass, distillation on $2500$ labels yields a difference in correlation of $0.11$ ($\mathrm{UD}_{2500}$ vs $\mathrm{D}_{2500}$); training with more data labels can obtain a similar value ($0.420$ vs $0.421$, $\mathrm{D}_{2500}$ vs $\mathrm{UD}_{10000}$). This value, however, increases to an even higher value ($0.69$; $\mathrm{D}_{10000}$) with distillation.

\subsection{Ivis Result}

\begin{figure}[ht]
\centering
  \includegraphics[width=8cm,height=2cm]{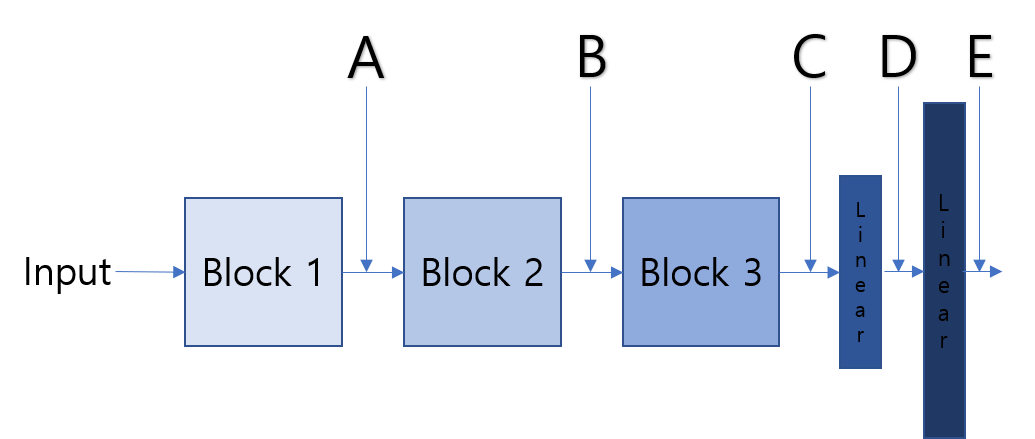}
  \caption{Extraction point of the embeddings}
\label{extractionpoint}

\end{figure}
  Next, we used the Ivis embeddings to analyze the effects of distillation on each model's layer-wise outputs. Fig.\ref{extractionpoint} shows the five different positions at which we calculated such embeddings. Table \ref{table_loss} shows the loss at convergence from the Ivis framework. 
  
\begin{table}[ht]

    \caption{Ivis convergence loss at the different output stages (Fig. \ref{extractionpoint}). The error across five folds is given as max-min range and was similar across undistilled (UD) and distilled (D) networks.}
    \centering
    \setlength{\tabcolsep}{5pt}
    \begin{tabular}{l|cc|cc|cc|cc|cc}
    \toprule

        {} & \multicolumn{2}{c|}{A ($\pm0.121$)} 
        &\multicolumn{2}{c|}{B ($\pm0.071$)} &\multicolumn{2}{c|}{C ($\pm0.024$)} &\multicolumn{2}{c|}{D ($\pm0.015$)}
        &\multicolumn{2}{c}{E ($\pm0.016$)}\\
    \cmidrule(lr){2-3} \cmidrule(lr){4-5} \cmidrule(lr){6-7} \cmidrule(lr){8-9} \cmidrule(lr){10-11}
    Labels      & UD & D  & UD & D & UD & D & U  & UD & UD  & D \\
    \midrule
    
    400   & 0.255 & 0.606 & 0.360 & 0.513 & 0.229 & 0.309 & 0.181 & 0.240 & 0.146 & 0.174 \\
    2500  & 0.274 & 0.533 & 0.375 & 0.530 & 0.380 & 0.312 & 0.317 & 0.246 & 0.240 & 0.180 \\
    5000  & 0.442 & 0.571 & 0.473 & 0.507 & 0.392 & 0.323 & 0.297 & 0.243 & 0.220 & 0.168 \\
    10000 & 0.462 & 0.498 & 0.474 & 0.515 & 0.391 & 0.305 & 0.289 & 0.227 & 0.204 & 0.156 \\
    20000 & 0.436 & 0.472 & 0.535 & 0.520 & 0.260 & 0.299 & 0.249 & 0.215 & 0.176 & 0.154
    \label{table_loss}
    \end{tabular}
\end{table}

We did not observe any \textit{significant} pattern of distillation at early positions A and B. However, the total variability in loss decreased as the embedding moved closer to the final layers from $\pm0.121$ to $\pm0.016$, showing that the embeddings created in early layers had notable fluctuation with different folds. Indeed, we observed that with very few labels (400), the loss for distilled network embeddings was always considerably higher - this should be taken with caution, however, given that the undistilled network may not produce reliable embeddings with only four images per class to begin with.

Interestingly, we also observed few significant differences in distilled networks as a function of label budget. This is most likely due to using the additional number of unlabeled data for extra training. In contrast, more labels made the loss of the undistilled network slowly follow its respective distilled network. Most importantly, we also found that the distilled network had reduced loss at the later C, D, and E layers with more than 2500 labels compared to its undistilled cousin.

Overall, we were able to detect several effects of distillation on the converging loss. Since the embeddings were independent of the test output of each classifier model, the only cause for such different losses would be due to differences in the raw representation space of the test dataset. Based on the overall loss criteria of the Ivis model, we suggest that the distilled network's lower layers has increased potential to separate negative and move closer positive pairs, resulting in "simpler" embeddings especially in low label-budget conditions.
 
\subsection{Quantitative analysis of representation space}
  
\begin{figure}[ht]
    \centering
     \includegraphics[width=\textwidth]{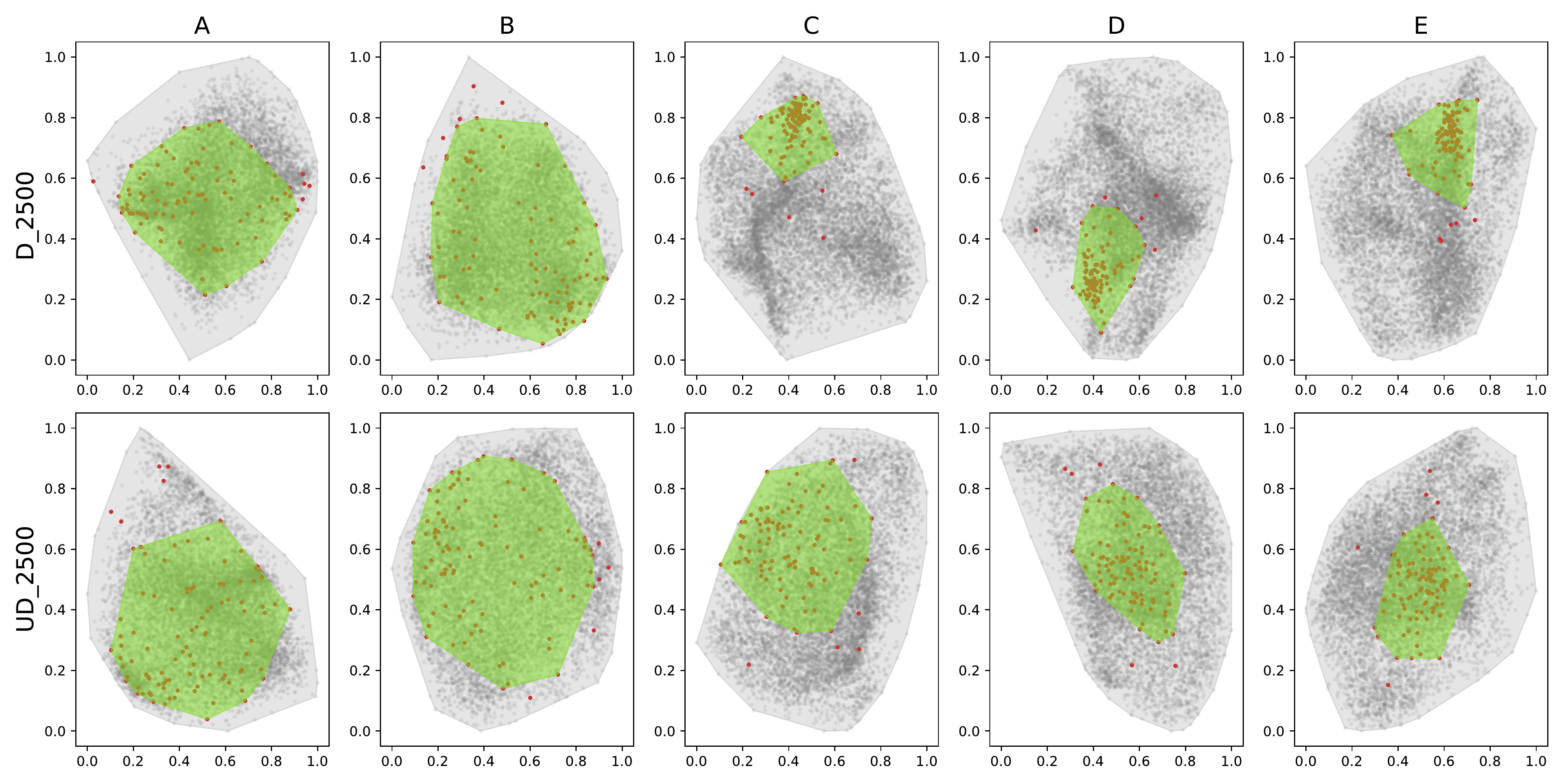}
    \captionof{figure}{Single class("plate") distribution of whole data}
    \label{fig:visualization_dis}
    
\end{figure}

As an additional, quantitative test we computed the boundary of a single class (plate) in CIFAR-100 over the whole test representation space using Gaussian density estimation (see  Fig.\ref{fig:visualization_dis}). Here, we can see that the distilled network has earlier, tight clustering of this class in its representation space, confirming our earlier analyses of correlations above.

\begin{figure}[ht]

\centering
  \includegraphics[width=\textwidth]{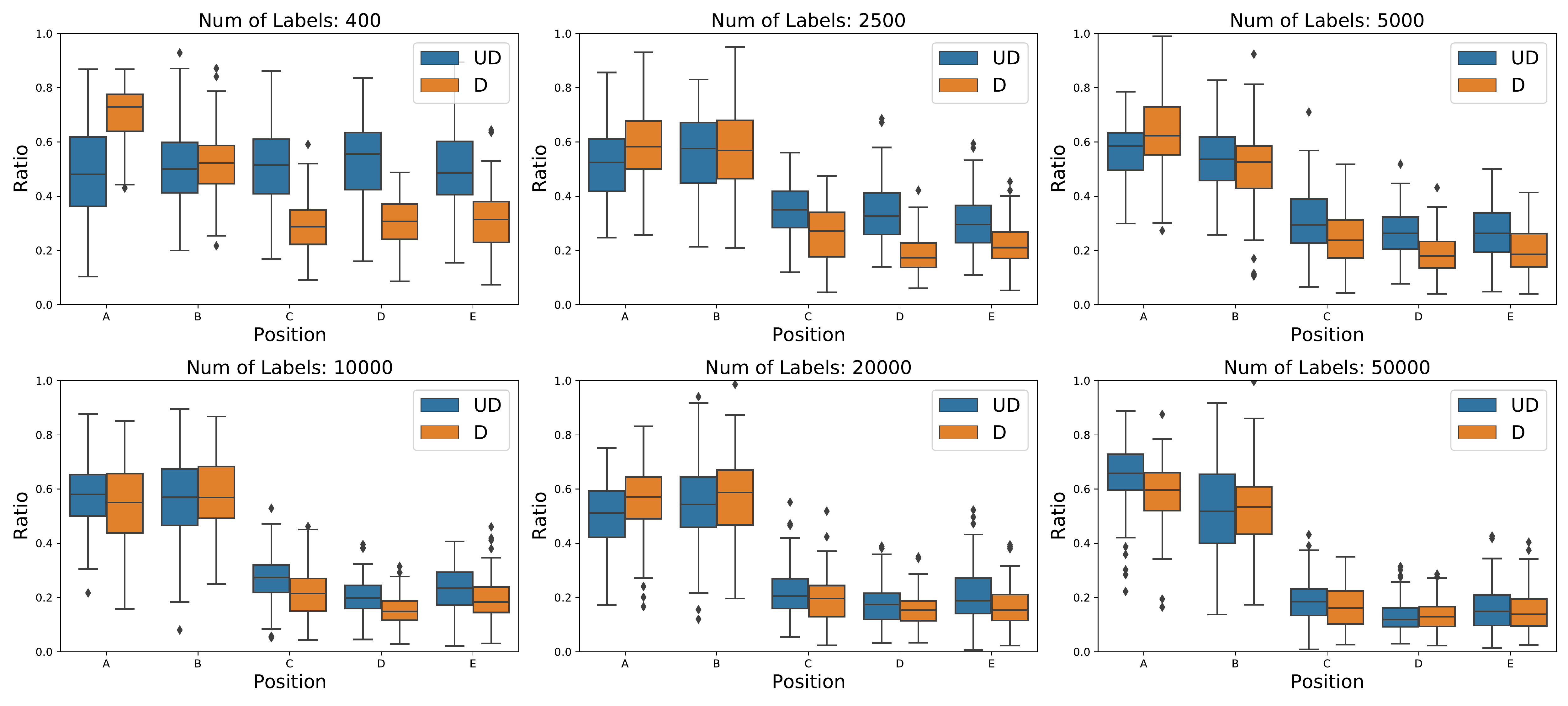}
\captionof{figure}{Box plot of area ratio for all classes}
\label{fig:perimeter Ratio}

\end{figure}

Fig.\ref{fig:perimeter Ratio} shows the computation of the mean class area to the whole area for different label budgets across extraction points (layers). There was no change in this ratio across layers for the undistilled network in the 400 label case, as the network was not able to distribute the classes into different chunks. However, even here the distilled network already showed crucial differences between the early and the late layers of the network. This difference was visible for all label budgets, separating the early layers A and B from the later layers C-E.  In addition, with increasing label budget, the gap between the two networks was slowly reduced across layers, matching our earlier loss analysis. Again, this analysis confirms that the distilled network finds a more tight (potentially effective) representation faster (i.e., in earlier layers) in the network for low label budgets.

\section{Conclusion and Future Work}

    All our results clearly showed how distillation provides the network with increased robustness at low label budgets. Crucially, we verified this not only in terms of accuracy, but also with our detailed loss analysis and class measurements using the Ivis method. This increased robustness comes in part due to the additional provision of unlabeled data, which represents one advantage of the distillation framework given the abundance of unlabeled data in the wild.
    
   We also showed that Ivis is able to visualize representation spaces even in relatively high-dimensional layers. With this visualization, we observed more compact class representations in distilled networks in general, happening at earlier layers. We presume that the teacher network's pseudo-outputs provide large amounts of feature information feedback to the student, leading the distilled network to make every layer work more independently and effectively. In our case, the first block in the Wide-Resnet has 16 filters per inner convolution layers, followed by 32 and 64 filters later. The best possible outcome would occur when every filter detects different features - an argument made in \cite{huh2021low}, which showed that the performance of deeper networks is better due to the kernels capturing more independent information. We argue that distillation also works in a similar fashion, resulting in better performance. In addition, our preliminary comparisons with t-SNE indicated that Ivis may be better suited for such high-dimensional visualizations, but further studies also on a larger variety of datasets will be necessary to validate this observation.  

 In our results, we also observed a large degree of similarity between the recent FixMatch approach and distillation training with a similar model architecture on CIFAR-100 training - future work will need to analyze different model architectures with better, overall performance in more detail. FixMatch's main idea was to match the distribution of output with soft augmented and hard augmented images without damaging the essential features of the input source - distillation is actually similar to this approach as the large network's output preserves the important feature information. One limitation of FixMatch compared to distillation is that the augmentation methods of the former need to be hand-crafted, which varies from dataset to dataset (faces vs objects, for example). We cautiously suggest therefore that distillation overall has the better potential to generalize to other dataset domains, but this remains to be shown in future work as well.  

\section{Acknowledgments}

This work was supported by Institute of Information Communications Technology Planning Evaluation (IITP) grant funded by the Korean government (MSIT) (No. 2019-0-00079), Department of Artificial Intelligence, Korea University

\bibliographystyle{splncs04}

{\small
\bibliography{paper_v0}
}

\end{document}